# Mapping Industry 4.0 Technologies: From Cyber-Physical Systems to Artificial Intelligence

Benjamin Meindl[1,*] & Joana Mendonça[1,2]

[1] Center for Innovation, Technology and Policy Research (IN+), Instituto Superior Técnico, Universidade de Lisboa, Avenida Rovisco Pais, 1049 - 001 Lisboa, Portugal
[2] CEiiA, Avenida Dom Afonso Henriques 1825, Matosinhos, Portugal

**ABSTRACT**

The fourth industrial revolution is rapidly changing the manufacturing landscape. Due to the growing research and fast evolution in this field, no clear definitions of these concepts yet exist. This work provides a clear description of technological trends and gaps. We introduce a novel method to create a map of Industry 4.0 technologies, using natural language processing to extract technology terms from 14,667 research articles and applying network analysis. We identified eight clusters of Industry 4.0 technologies, which served as the basis for our analysis. Our results show that Industrial Internet of Things (IIoT) technologies have become the center of the Industry 4.0 technology map. This is in line with the initial definitions of Industry 4.0, which centered on IIoT. Given the recent growth in the importance of artificial intelligence (AI), we suggest accounting for AI's fundamental role in Industry 4.0 and understanding the fourth industrial revolution as an AI-powered natural collaboration between humans and machines. This article introduces a novel approach for literature reviews, and the results highlight trends and research gaps to guide future work and help these actors reap the benefits of digital transformations.

## 1. INTRODUCTION

The fourth Industrial revolution, often referred to as Industry 4.0, describes the integration of emerging technologies and digital tools across the manufacturing value chain, building on cyber-physical systems (CPSs). The term was coined in 2011 by the Industrie 4.0 working group, which produced the report to *Secur[e] the future of [the] German manufacturing industry* (Kagermann, Wahlster, & Helbig, 2013, p. 17). The concept of Industry 4.0 addresses the significant transformations in various industrial sectors (Roblek et al., 2016) that have created the opportunity

---

[*] Corresponding author: Benjamin Meindl, Avenida Rovisco Pais, 1049 - 001 Lisboa, Portugal

Contact: benjamin.meindl@tecnico.ulisboa.pt

to reshape manufacturing (Krzywdzinski et al., 2016) and enable sustainable development (Bai, Dallasega, Orzes, & Sarkis, 2020). Aside from the German task force, several other national initiatives have introduced ideas for how to address the technological changes of the fourth industrial revolution (Santos, Mehrsai, Barros, Araújo, & Ares, 2017)—for example, "Made in China 2025" (Li, 2018), the New Industrial France program (NIF, 2016), the U.K. Foresight project (Foresight, 2013), and the U.S. Smart Manufacturing Operations Planning and Control Program (NIST, 2014).

Following these initiatives, an increasing amount of research has focused on manufacturing in the context of the fourth industrial revolution. However, a precise characterization of the technological scope of Industry 4.0 is lacking (Zheng, Ardolino, Bacchetti, & Perona, 2020). Rüßmann et al. (2015) described, for example, nine high-level technology drivers (core technologies), including autonomous robots and cybersecurity. Y. Chen (2017) presented 10 major technologies, including 3D printing and virtual reality. Culot, Nassimbeni, Orzes, and Sartor (2020) used a framework of 13 enabling technologies, while others (Beier, Ullrich, Niehoff, Reißig, & Habich, 2020) described 15 key technology features. Both Culot et al. (2020) and Beier et al. (2020) reviewed defining elements of previous Industry 4.0 concepts by comparing their frameworks against previous publications. Culot et al. (2020) found, for example, that only one Industry 4.0 article referred to "new materials" as a key enabling technology, whereas the Internet of Things (IoT) was a key technology in many articles.

Industry 4.0 is characterized not only by certain technologies, but also by their connections and combinations (Alcácer & Cruz-Machado, 2019; Bai et al., 2020; Culot et al., 2020). Analyzing the relations between technologies complements previous work and provides a better understanding of the evolution of the Industry 4.0 landscape (Meindl, Ayala, Mendonça, & Frank, 2021). We build on Chiarello, Trivelli, Bonaccorsi, and Fantoni's (2018) approach to create a network of Industry 4.0 technologies. Building a technology map shows the relevance of technologies, technology clusters, and their interrelations. Whereas Chiarello et al. (2018) based their work on an analysis of Wikipedia articles, we used natural language processing (NLP) to extract information on technologies from scientific publications. Our work differs from many previous review articles in that it does not rely on a review of conceptual articles but rather builds on the extensive amount of articles about Industry 4.0, including those focusing on specific aspects of the concept. This search yielded more than 14,000 articles published over a period of 10 years. Further, we introduce a timeline in our analysis, which allows us to review the evolution of the concept since 2011. Finally, the obtained technology map provides an extensive dataset of underlying technologies and their relations to each other.

This article makes two main contributions. First, the analysis shows the evolution and trends of Industry 4.0 technologies, and helps identify further avenues for research to ensure progress in the transformation towards Industry 4.0. As Industry 4.0 is a relatively new, fast-changing concept (Culot et al., 2020; Galati & Bigliardi, 2019), we provide a dynamic view of Industry 4.0 that both provides a snapshot of the technological landscape and shows historical development, enabling it to be updated and to track its progress in the future. Unlike previous articles, we do not rely on predefined Industry 4.0 dimensions. Instead, our approach uses network analysis to create an unbiased bottom-up view of the Industry 4.0 landscape. We identify eight technology clusters with Industry 4.0 at the center. Our trend analysis shows that the importance of artificial intelligence (AI) has seen major growth. We suggest future frameworks to reflect this trend and emphasize the important role of AI in Industry 4.0. Second, this article provides a novel approach for a systematic literature review. The method extracts relevant information from the Industry 4.0 academic literature using NLP, and we develop a mapping algorithm to create a technology map for visualization and network analysis. The proposed methodology may also be useful in reviewing the evolution of other research fields.

The remainder of the paper is structured as follows. The next section presents a review of the literature on Industry 4.0 technologies, after which Section 3 presents the approach used. The results (Section 4) include an Industry 4.0 technology map and the evolution of technology clusters over time. A discussion of the results is presented in Section 5, and the conclusion follows in Section 6.

## 2. INDUSTRY 4.0 IN THE LITERATURE

### 2.1. Definitions of Industry 4.0

In their initial article, the Industrie 4.0 working group described Industry 4.0 as a broad industrial paradigm (Kagermann et al., 2013). The smart factory and CPSs form the core of this idea, and interact with a smart environment, including smart mobility, smart logistics, smart buildings, smart products, and smart grids. The U.K. Foresight program (Foresight, 2013) and the NIST Smart Manufacturing concepts (NIST, 2014) have a strong focus on manufacturing, as well, whereas the New Industrial France program (NIF, 2016) has a broader focus that includes eco-mobility and smart cities as core focus areas. The Chinese Manufacturing 2025 program has a particular focus on AI (Li, 2018; Zhou et al., 2018).

Several research articles have focused on the core of Industry 4.0, which the Industrie 4.0 working group (Kagermann et al., 2013) report describes as the "Smart Factory" (Meindl et al., 2021). A framework developed by Boston Consulting Group (Rüßmann et al., 2015) has frequently served as a reference point for research (Alcácer & Cruz-Machado, 2019; Butt, 2020; Saucedo-Martínez, Pérez-Lara, Marmolejo-Saucedo, Salais-Fierro, & Vasant, 2018). This concept of Industry 4.0 comprises nine technologies, with a narrow focus on industrial production plants. These technologies include, for example, autonomous robots, augmented reality, and additive manufacturing. Y. Chen (2017) introduced the concept of "integrated and intelligent manufacturing," which incorporates similar technologies as the concept developed by Rüßmann et al. (2015) but focuses on software platforms integrating these technologies.

4H. Lu and Weng (2018) reviewed 30 international case studies to evaluate technology maturity and roadmaps. They also focused on the smart factory aspect of Industry 4.0. The article structured the technologies in four layers to represent a factory infrastructure: An integration layer connects the sensor layer to an intelligent layer that conducts analytics, and the response layer uses information from the intelligent layer to generate actions, such as production and sales prediction. Similarly, other authors have described an Industry 4.0 landscape tightly centered around the smart factory (Brettel, Friederichsen, Keller, & Rosenberg, 2014; Dalenogare, Benitez, Ayala, & Frank, 2018). These articles emphasize the integration of flexible shop floor production and product development processes as a core idea in Industry 4.0. Zhong et al. (2017) described a similar vision, termed "intelligent manufacturing." Their article described intelligent manufacturing as a synonym of smart manufacturing and part of a broader Industry 4.0 paradigm. Bai et al. (2020) also highlighted the role of intelligent systems, describing Industry 4.0 as "a new paradigm of smart and autonomous manufacturing" (p. 2). Unlike most other frameworks, they also considered drones and global positioning systems as key technologies.

Lasi et al. (2014) drew a broader picture of Industry 4.0 technologies. In addition to the factory itself and the development and monitoring processes, their Industry 4.0 vision focuses on logistics, including delivery and suppliers of goods and tools. Their concept thus focuses on a cyber-physical production network, including innovative enterprise resource planning (ERP) systems. Ghobakhloo (2018) presented a framework focusing not only on the smart factory but also on the integration of logistics and customers (e.g., the Internet of People) as key elements of the Industry 4.0 landscape.

A. G. Frank, Dalenogare, et al. (2019) combined many of these ideas into a framework that describes the Industry 4.0 enterprise as consisting of four "smart" elements. Aside from smart manufacturing (the smart factory), Industry 4.0 comprises a smart supply chain (e.g., platforms with suppliers), smart working (e.g., augmented reality for product development), and smart products (product connectivity). In addition to some underlying base technologies, such as the IoT, they argue that each of the "smarts" is related to different technologies. Nakayama, de Mesquita Spínola, and Silva (2020) also highlighted the relevance of Industry 4.0 beyond company boundaries and emphasized the flexibility of decentralized systems as the differentiator from Industry 3.0. Similarly, Alcácer and Cruz-Machado (2019) emphasized the role of Industry 4.0 in decentralization and real-time engagement. Roblek, Meško, and Krapež (2016) described Industry 4.0 at a higher level: Aside from the enterprise and supplier level, they also described smart cities and digital sustainability as fundamental components of Industry 4.0.

*2.2.   Reviews of Industry 4.0 concepts in the literature*

Previous review articles on Industry 4.0 indicate how the concept has been perceived in academia and industry. Table 1 summarizes the various literature review approaches to date.

Ghobakhloo (2018) reviewed Industry 4.0 design principles and technology trends in an article that used language processing tools to review 178 articles and book chapters and described Industry 4.0 as a broad concept reaching far beyond the smart factory. Ghobakhloo (2018) identified technology trends—including blockchain, additive manufacturing, and the Internet of People—as well as Industry 4.0 design principles, which include decentralization, integration, individualization, and smart entities. Similarly, in a manual review of 88 articles, Yang Lu (2017) derived a broad concept of Industry 4.0 from the literature, including the smart city, smart grid, and smart home as applications of Industry 4.0. Another manual review was conducted by Saucedo-Martínez et al. (2018), who reviewed 110 articles and provided information on their contributions along with nine key technologies defined by Rüßmann et al. (2015). They found that horizontal and vertical integration across company boundaries was the pillar of Industry 4.0.

| Literature review | Articles reviewed | Method |
|---|---|---|
| Chiarello et al. (2018) | Wikipedia articles related to Industry 4.0 | Automated crawling |
| Culot et al. (2020) | 81 academic and 18 non-academic sources | Manual |
| Ghobakhloo (2018) | 178 academic and non-academic articles and book chapters | IBM language processing and manual review |
| Hermann, Pentek, and Otto (2016) | 130 academic and non-academic sources | Natural language processing |
| Kipper, Furstenau, Hoppe, Frozza, and Iepsen (2020) | 1,882 academic articles | Keyword analysis |
| Liao, Deschamps, Loures, and Ramos (2017) | 224 academic articles | Keyword analysis |
| Muhuri, Shukla, and Abraham (2019) | 1,619 academic articles (focusing on most-cited articles) | Bibliometric review |
| Saucedo-Martínez et al. (2018) | 110 academic articles | Manual |
| Strozzi, Colicchia, Creazza, and Noè (2017) | 462 academic articles | Bibliometric review |
| Yang Lu (2017) | 88 academic articles | Manual |
| Zheng et al. (2020) | 186 academic articles | Manual |

*Table 1: Overview of Industry 4.0 technology literature reviews.*

Unlike the previously described reviews, Liao et al. (2017) undertook a quantitative analysis. They found that Industry 4.0 articles referred most commonly to the manufacturing stage of the product lifecycle. Most common technology keywords referred to information technology, such as modeling, visualization, or big data. Their systematic keyword analysis built on 224 academic articles. Similarly, Hermann et al. (2016) conducted a quantitative text analysis of 130 publications related to Industry 4.0, including scientific articles as well as practical journals and books. They used NLP to create an overview of the most frequent terms in publications and used insights from an expert workshop to identify four Industry 4.0 design principles: technical assistance (e.g., virtual assistance); interconnection (e.g., standards); information transparency (e.g., data analytics); and decentralized decisions. Hermann et al. (2016) provided a vision of Industry 4.0 that centered on the smart factory.

Zheng et al. (2020) conducted a manual review of 186 selected Industry 4.0 articles structured along the business process life cycle. They identified Industry 4.0 applications such as order picking management, collaborative operations with humans, and assembly defect detection. They linked these applications to core technologies, such as IoT, AI, and additive manufacturing. Culot et al. (2020) systematically reviewed 81 academic and 18 non-academic articles on Industry 4.0, addressing the enabling technologies, characteristics, and expected outcomes related to Industry 4.0. Unlike previous reviews, they not only focused on Industry 4.0 but also included different conceptual terms, such as smart manufacturing and cyber manufacturing. They described Industry 4.0 as a broad concept that covers various more specific concepts, such as cloud manufacturing, which focuses on new business models, and social manufacturing, which includes society and consumers as key characteristics. This differentiation among concepts represents a novel contribution to the literature. Like most reviews, Zheng et al. (2020) and Culot et al. (2020) mainly built on articles that aimed to provide a conceptual view of Industry 4.0. Since our work incorporates a broader range of articles on Industry 4.0, we included all articles that refer to Industry 4.0, including case studies and articles about specific technologies (e.g., 5G, 3D printing) or specific contexts (e.g., the supply chain). This allowed us to reflect not only on insights from the Industry 4.0 research community but also from other fields, such as computer science.

Due to the large number of articles referring to Industry 4.0 and related topics, a truly comprehensive manual review is difficult to achieve. Therefore, research may make use of automated methods. For example, Strozzi et al. (2017) and Muhuri et al. (2019) conducted bibliometric reviews of the smart factory and Industry 4.0 literature using structured textual data. Although they mentioned the most frequent keywords and keyword clusters, their main focus was identifying citation networks and the most relevant institutions and authors. Chiarello et al. (2018) reviewed a large number of articles using an automated review to create a network map of Industry 4.0 technologies that provides information on the importance of technology fields, as well as clusters and their interrelations. Their analysis used seed words from scientific articles and industry reports to crawl Wikipedia articles related to Industry 4.0 for further keywords. This approach identified more than 1,200 technologies as an input for their analysis, most of them related to information technologies. The analysis described 11 technology clusters, including big data, programming languages, and computing. Some researchers have identified as a limitation the lack of an agreed-upon Industry 4.0 taxonomy that can be used for a structured review (Zheng et al., 2020). This barrier can be overcome when building a technology network, which relies on articles themselves as a basis for technology clusters or categories, shows the relations between clusters, and is independent of previous frameworks (see Chiarello et al. 2018).

## 3. METHODS

We aimed to contribute a better understanding of the Industry 4.0 technology landscape, its evolution, its key technologies, and the relations among those technologies. Therefore, we developed an automated approach to review the literature. First, we retrieved relevant articles from Scopus using its API and removed duplicate articles and retracted articles. Second, we trained a neural network to identify technology terms in the article abstracts, which we then extracted. Finally, we visualized Industry 4.0 technologies as a technology map and conducted network analysis. The results show connections between technologies, as well as results from network analysis that can better explain the importance of technologies, clusters and trends. Fig. 1 illustrates the flow of the analysis.

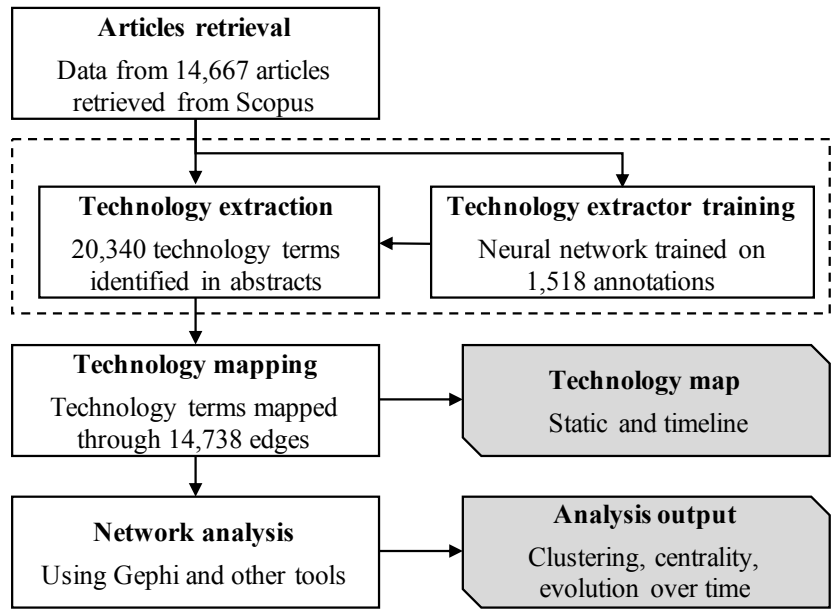

*Fig. 1.* Flow diagram representing our analysis. First, documents were extracted from Scopus and prepared for analysis. Next, keywords were extracted and mapped, and finally network analysis and visualization were used to show clusters and trends of technologies. White and grey boxes indicate analysis processes and results, respectively. [1-column]

### 3.1. Data

Our analysis used scientific publication data collected from Scopus, one of the largest repositories of scientific articles (Harzing & Alakangas, 2016; Liao et al., 2017) and a reliable source that has been used in previous reviews (Kipper et al., 2020; Meindl et al., 2021; Pirola, Boucher, Wiesner, & Pezzotta, 2020). In addition to article titles, abstracts, and keywords, we retrieved the first author's country to analyze the geographical distribution of research conducted.

Industry 4.0 technologies are described with various conceptual terms (Culot et al., 2020); therefore, we created a search query with the most important related terms. Industry 4.0–related terms include: "Industry 4.0" (Chiarello et al., 2018; Y. Lu, 2017); "Industrie 4.0" (Kagermann et al., 2013); "digital manufacturing" (Behrendt, Kadocsa, Kelly, & Schirmers, 2017; Chryssolouris et al., 2009; Paritala, Manchikatla, & Yarlagadda, 2017); "smart manufacturing" (Kang et al., 2016; Kusiak, 2018; Tao, Qi, Liu, & Kusiak, 2018); "intelligent manufacturing" (Zhong et al., 2017); "cloud(-based) manufacturing" (Tao, Cheng, Xu, Zhang, & Li, 2014; Wu, Greer, Rosen, & Schaefer, 2013); "Factory/Factories of the Future" (European Commission, 2013); and "advanced manufacturing" (Cheng et al., 2018; R. Shah, 1983). We included articles available in Scopus on March 30 2020, published since 2011 that contained any of these terms in their title, abstract, or keywords. The search yielded 14,667 articles. Fig. 2 presents the number of papers per year per term and shows that the number of papers related to Industry 4.0 and smart manufacturing increased significantly since 2014. Since 2016, the search term "Industry 4.0" has generated the most results.

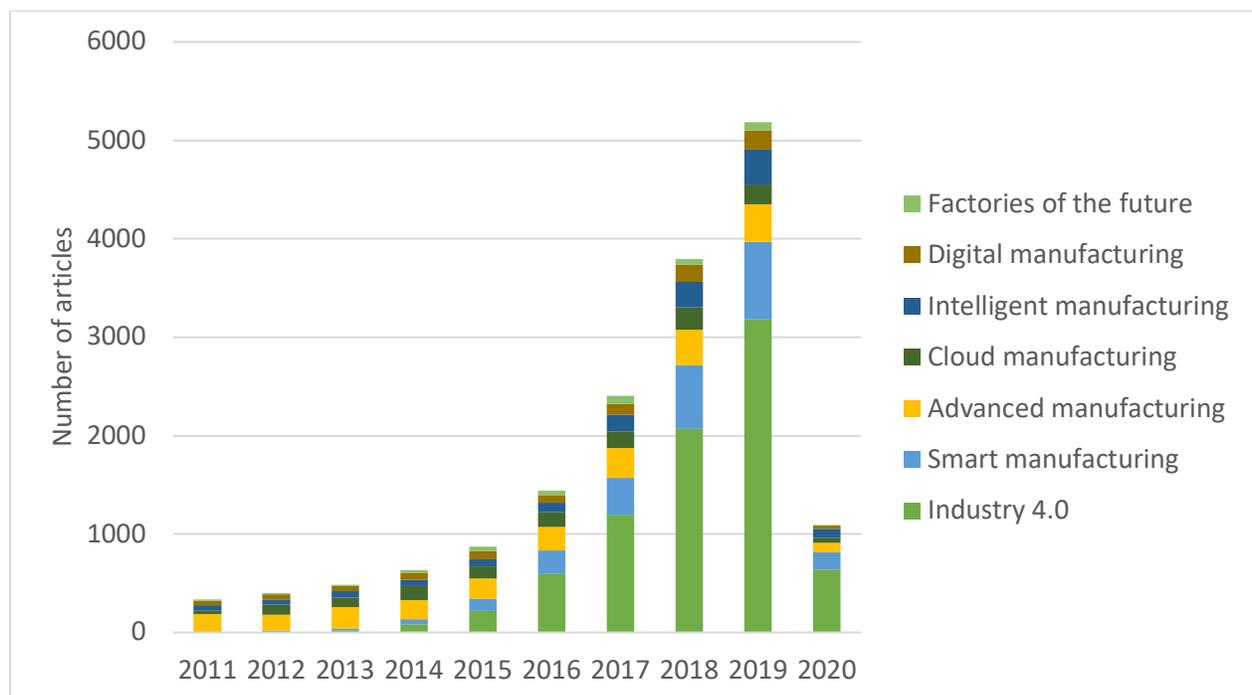

*Fig. 2.* *Distribution of papers associated with search terms. Multiple search terms per paper are possible. Terms are ordered by total growth in number of articles between 2011 and 2019. [1-column]*

We used publications' abstracts for our analysis. Abstracts are particularly useful because they provide the highest ratio of keywords to text (P. K. Shah, Perez-Iratxeta, Bork, & Andrade, 2003).

Similar analyses based on journal abstracts have been conducted to (for example) identify technological fronts (Shibata, Kajikawa, Takeda, & Matsushima, 2008) or obtain insights into applications of machine learning in smart manufacturing (Sharp, Ak, & Hedberg, 2018).

*3.2.   Technology terms extraction*

We used NLP to extract technology terms from documents. Extracting specific entities from a text—such as a location, person, or (in this case) technology term—is called named entity recognition (NER). NER can rely on handcrafted rules for feature engineering (Lample, Ballesteros, Subramanian, Kawakami, & Dyer, 2016), such as searching for combinations of adjectives and nouns. Analyzing word frequencies (Shibata et al., 2008) has also commonly been used to identify keywords. Zhang, Porter, Hu, Guo, and Newman (2014), for example, used a word-frequency–based method called tf–idf to identify keywords related to dye-sensitized solar cells, and Prabhakaran, Lathabai, and Changat (2015) relied on this method to identify paradigm shifts in emerging information technology fields.

This work takes an approach based on neural networks refined with feature engineering rules. NER methods based on neural networks are superior to other approaches in many cases (Yadav & Bethard, 2019), as they can achieve better accuracy and require less intensive feature engineering. A neural network learns which terms to extract based on manually annotated examples. The algorithm identifies terms not only by remembering words but also using neighboring words. This method enables the algorithm to identify technology terms that have not previously been manually annotated.

This article relies on the spaCy neural network model (Honnibal & Montani, 2017), which we train based on 1,518 manual annotations. Additionally, 454 annotations were used to evaluate the model, which reached an accuracy (precision) rate of 78%, indicating that most technologies were correctly identified[†]. This score exceeds previous approaches, such as that of Anick et al. (2014), who reached a precision rate of 63% for English technology terms based on 3,700 manual annotations when extracting technology terms from patents. To account for false positives, we included manual cleaning steps (e.g., removing frequent falsely identified non-technology terms). Further, since many technology terms consist of multiple words, we introduced an approach particularly suitable for identifying those terms. For example, instead of "flexible and reconfigurable manufacturing systems," our method would identify both "flexible manufacturing systems" and "reconfigurable manufacturing systems." This was achieved by training the algorithm to identify only the head word (in this case, "systems") and then using sentence parsing to extract the technology terms. The Appendix provides a detailed description of this approach.

---

[†] The model was repeatedly trained during annotating. Annotating was stopped when additional annotations did not lead to significant improvements in accuracy.

After these steps, a few non-technology terms may still not be filtered out, particularly low-frequency terms. We did not expect a structured bias to result from those words, as they appeared across all types of articles. Therefore, we expected that occasional non-technology terms would not have a significant negative impact on the network but would add some noise.

*3.3. Mapping technology terms*

Our neural network extracted technology terms from each abstract in our analysis. The novel mapping algorithm mapped technologies based on co-occurrence and semantic similarity. Additionally, the document's timestamp was added to each term to allow for an analysis of evolution over time.

Network analysis allows insights from large amounts of data to be connected and has been used to take advantage of the increased availability of data. Kim, Lee, and Kwak (2017) used patent network data to research the influence of standards on the convergence of IoT technology. Further, Kwon and Park (2015) proposed a framework that generates social impact scenarios of new technologies based on text mining of online content, while another group of researchers used network analysis to analyze the Dutch innovation system (van Rijnsoever, van Den Berg, Koch, & Hekkert, 2015). Network analysis has also been conducted to elucidate Industry 4.0 (e.g., Chiarello et al. 2018; Muhuri et al. 2019). Meindl et al. (2021) used network analysis to visualize the integration of different aspects of Industry 4.0 over time.

We mapped technology terms based on co-occurrence, which indicates a relatedness between technologies (Chiarello et al., 2018; Shibata et al., 2008; Yoon & Kim, 2012; Zhang et al., 2014). Technologies mentioned in the same article abstract were considered linked. Technologies co-occurring in the same sentence received an additional link, as this indicates close relatedness. The weight of all links within a document adds up to one, ensuring the equal influence of each article on the overall technology map. Additionally, we conducted semantic mapping to ensure that sub-technologies were linked to their base technology. For example, "wireless sensor network" is linked to "wireless sensor." Therefore, we defined the strength of the semantic link between two words such that on average the sum of semantic links per technology term equaled those based on co-occurrence. This ensured that network weights were still mainly driven by technology term frequencies while still reflecting semantic relations.

*3.4. Network visualization and analysis*

The nodes (technology terms) and edges (links between terms) were imported into the tool Gephi (Bastian, Heymann, & Jacomy, 2009) for visualization and network analysis. The nodes were arranged for visualization using the ForceAtlas2 algorithm (Jacomy, Venturini, Heymann, & Bastian, 2014), which allows for high-quality, intuitive mapping. The algorithm arranges the technologies in a force-directed layout where nodes repulse each other while edges (e.g., co-occurrence) act like springs, attracting nodes.

To analyze the importance of a single technology within the technology map, we calculated two measures: the weighted degree and the eigenvector centrality. The degree describes a node's size. The weighted node degree accounts for the weights of node connections in weighted networks and is defined as the sum of the weights of all connections linked to a node (Opsahl, Agneessens, & Skvoretz, 2010). The weighted degree only accounts for a node's local network (direct connections) and thus does not well represent its importance for the overall network. Therefore, we introduced eigenvector centrality as a second measure. The eigenvector score is higher for nodes connected to other nodes with a high eigenvector (Bonacich, 2007). For the analysis, we used the measures as follows: Weighted node degree served as a filter to select the technologies most frequently mentioned in literature, while eigenvector centrality served as an additional measure for importance—for example, to analyze how the importance of a technology changed over time.

Further, we identified technology clusters. Gephi offers modularity analysis, which divides a network into clusters of closely related nodes (Blondel, Guillaume, & Lefebvre, 2008). We conducted modularity analysis using Gephi and assigned different colors to the technologies in each cluster (Lambiotte, Delvenne, & Barahona, 2009). Based on these clusters, we evaluated the relations between clusters. For each technology, we counted the number of connections to technologies in each cluster. Calculating the share of links from a technology to each cluster helps identify cluster-linking technologies: technologies that have strong connections with more than one cluster. In our analysis, we also considered that smaller clusters have, by definition, fewer incoming connections and therefore also normalized the shares of links per cluster by cluster size. Further, we calculated the strengths of connections between clusters by summing the connections of all nodes of a given cluster to the nodes of each remaining cluster.

## 4. RESULTS

In this section, we present the results of our network analysis. We show the overall technology network, clusters, and most important technologies. This analysis includes insights on cluster-bridging technologies and the evolution of the network over time.

Analysis of the scientific corpus resulted in 2,317 technology terms (modeled as nodes in the network). The nodes are connected via 14,560 edges. Cluster analysis identified eight technology clusters of strongly connected technologies, which we describe in detail below. On average, around 80% of a technology's connections occurred within its own cluster. These numbers suggest that clustering provided meaningful results, with clear cluster associations. The clusters represent broad fields of Industry 4.0, and we assigned names to each cluster to best represent its associated technologies. The network comprises a core IIoT cluster and seven outer clusters. IIoT marks the center of the technology map and is the largest cluster. It represents mainly technologies related to the IIoT and communication technologies. In addition, there are seven outer clusters: four related to information technology (Algorithms, Cloud Platforms, Management Systems, Sensor Systems), and three related to manufacturing processes (Additive Manufacturing, Computer-Aided Manufacturing [CAM], Human–Robot Systems). Fig. 3 visualizes the technology map, with colors indicating clusters. Additionally, the graph data is available for download and to explore online.‡

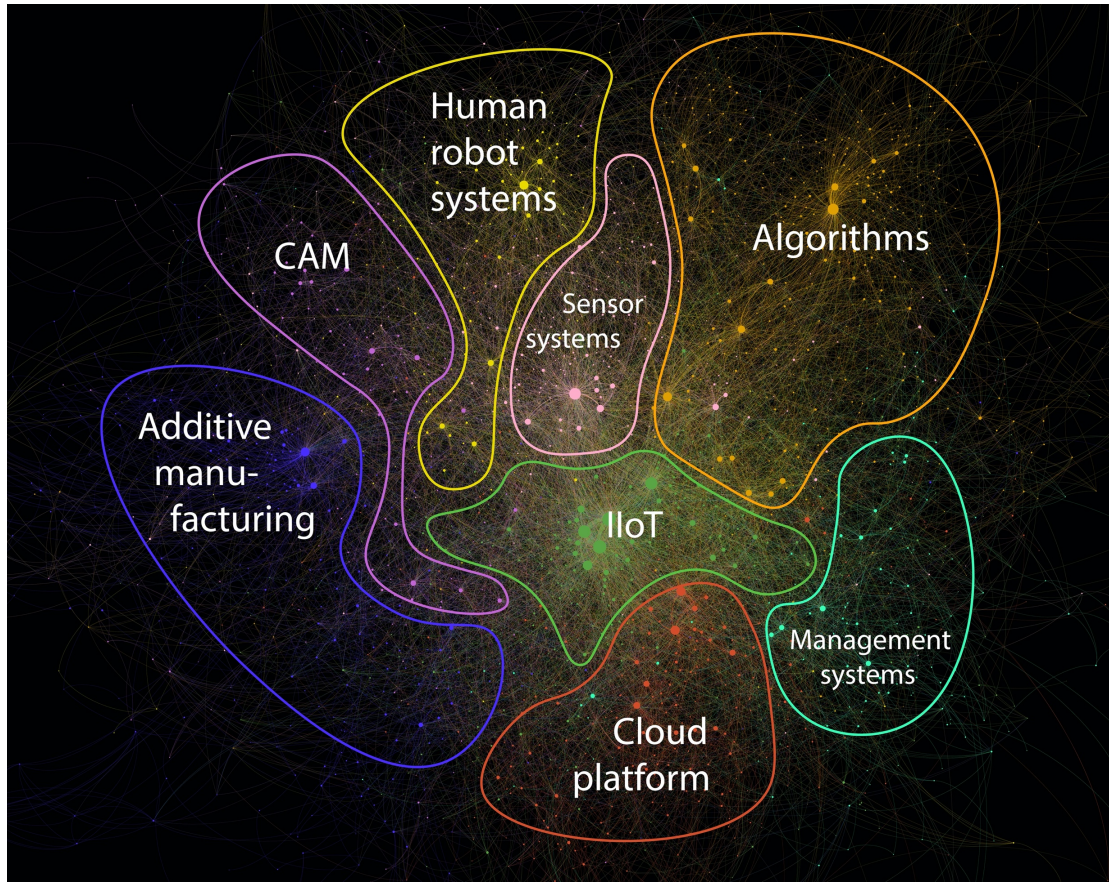

*Fig. 3. Map of technologies showing the eight technological clusters. Unassigned areas include technologies from various overlapping clusters. CAM stands for Computer-Aided Manufacturing and IIoT for Industrial Internet of Things. [1.5-column]*

Table 2 provides further information on technologies, such as centrality and degree. It includes the 10 highest-degree technologies of the technology network. We also included the three highest degree technologies per cluster, as well as some of the fastest-growing and fastest-declining technologies per cluster. Finally, the table includes relevant cluster-bridging technologies with strong connections to multiple clusters. The dynamic data allowed us to look into the development of technology clusters over time to identify trending technology fields. We used two measures for timeline analysis. First, Table 2 describes the change in importance of single technologies through a centrality measure (eigenvector centrality), which indicates how well connected (and thus relevant) a technology is within the overall network. Second, we described cluster importance

---

‡ https://bmeindl.github.io/technology_network/

13using the weighted degree of technologies related to the cluster. Fig. 4 illustrates the evolution of a cluster's importance in the last decade, measured as the sum of the weighted node degrees within a cluster as a share of the total weighted degree of the network. Fig. 5 includes additional information on cluster sizes and relations between clusters. The text below includes some information on other technologies not included in Table 2.

| Cluster<br>  Technology | EV centrality<br>(Δ 2018–2020) | Weighted degree | Related cluster |
|---|---|---|---|
| **IIoT: Industrial Internet of Things** | | | |
| Internet of Things (IoT) | 1 (0) | 2957 | |
| Cyber-physical system (CPS) | 0.84 (-0.03) | 2621 | |
| Internet | 0.82 (-0.04) | 1406 | (CP) |
| Industrial IoT (IIoT) | 0.47 (0.10) | 1365 | |
| Information and communication technology (ICT) | 0.31 (-0.05) | 252 | (CP) |
| IoT devices | 0.24 (0.05) | 266 | |
| Software-defined networking (SDN) | 0.16 (0.03) | 130 | CP |
| **SE: Sensor Systems** | | | |
| Sensor | 0.77 (-0.05) | 1558 | |
| RFID | 0.35 (-0.05) | 344 | (IIoT) |
| Wireless sensor network (WSN) | 0.34 (-0.04) | 342 | (IIoT) |
| Smart sensor | 0.26 (0.03) | 162 | (IIoT) |
| Automated guided vehicle (AGV) | 0.12 (0.02) | 102 | (HRS) |
| Embedded sensor | 0.08 (0.06) | 36 | (IIoT, AM) |
| **AL: Algorithms** | | | |
| Artificial intelligence | 0.52 (0.13) | 855 | HRS, (IIoT) |
| Algorithm | 0.32 (-0.02) | 763 | - |
| Machine learning | 0.30 (0.10) | 428 | - |
| Analytics | 0.19 (-0.06) | 115 | - |
| Blockchain | 0.18 (0.1) | 212 | - |
| Big data analytics | 0.17 (0.05) | 126 | IIoT, (CP) |
| Genetic algorithm | 0.15 (-0.06) | 290 | - |
| **MS: Management Systems** | | | |
| Manufacturing execution system (MES) | 0.26 (-0.04) | 338 | - |
| Enterprise resource planning (ERP) | 0.23 (-0.01) | 195 | - |
| Product lifecycle management (PLM) | 0.14 (0) | 174 | (CP) |
| Condition monitoring system (CMS) | 0.10 (-0.02) | 41 | CP |
| Supply chain management (SCM) | 0.09 (0.03) | 51 | (AL) |
| **CP: Cloud Platform** | | | |
| Cloud computing | 0.66 (-0.07) | 678 | IIoT, (AL) |
| Cloud | 0.28 (-0.08) | 257 | |
| OPC UA | 0.27 (-0.01) | 362 | CAM, (IIoT) |
| Simulation | 0.13 (0.02) | 83 | AM, HRS, (MS, AL) |
| Edge computing | 0.07 (0.04) | 24 | AL |
| Platform | 0.07 (0.03) | 48 | (HRS) |
| **AM: Additive Manufacturing** | | | |
| Additive manufacturing | 0.38 (0.04) | 1154 | - |
| 3D printing | 0.27 (0.08) | 393 | - |
| 3D printer | 0.17 (0.03) | 157 | CAM |
| Modeling | 0.05 (0.01) | 59 | CP |
| **CAM: Computer-Aided Manufacturing** | | | |
| Industrial control system (ICS) | 0.16 (0) | 100 | CP, (IIoT) |
| Machine tool | 0.16 (-0.03) | 211 | (HRS) |
| Control system | 0.11 (0.06) | 107 | - |
| Computerized numerical control (CNC) | 0.09 (-0.03) | 145 | - |
| Computer-aided design (CAD) | 0.05 (0.02) | 41 | MS, (AM, HRS) |
| **HRS: Human–Robot Systems** | | | |
| Robotics | 0.29 (0.09) | 273 | (AM) |
| Augmented reality | 0.26 (0.07) | 211 | AM |
| Robot | 0.21 (0.01) | 294 | - |
| Virtual reality | 0.19 (0.07) | 196 | AL |
| Autonomous robots | 0.15 (0.02) | 52 | AM, (SE, CP) |
| Collaborative robot/cobot | 0.12 (0.03) | 93 | (AL) |

*Table 2: This table includes the highest-degree technologies in the technology network and some of the fastest-growing technologies. The text refers to additional technology terms. EV means eigenvector, a measure of how central a node is to the overall network. The technologies are ordered by weighted degree, which indicates local importance (within a cluster), whereas EV centrality refers to a central role within the overall network. Information on related clusters includes strong and medium-strong (in parentheses) connections to other clusters.*

As illustrated in Fig. 4, the importance of the IIoT grew significantly in 2015, when it became the most important cluster. The cluster includes the overall most frequent technology terms in the network: IoT and CPS. Within this cluster, many IIoT-related terms increased in importance, including IIoT, (I)IoT device, and (I)IoT system. Some terms related to network infrastructure, such as software-defined networking, smart grid, and 5G, also experienced strong growth. Sensor systems overall grew moderately and have slightly declined since 2015–2016. Some of the fastest-growing technologies include smart or intelligent sensors, embedded sensors, human–machine interfaces, and predictive maintenance. The cluster also includes technologies that heavily rely on sensor networks, such as automated guided vehicles (AGVs) and autonomous vehicles, which have been increasing in importance over time.

The second most important cluster is Algorithms. After a decline between 2011–2012 and 2015–2016, this cluster has shown a slight growth in importance. Whereas terms such as genetic algorithm and multi-agent system used to be central within the Algorithms cluster, the recent growth in cluster importance is driven by the fast growth of AI, machine learning, blockchain, and big data analytics. AI and blockchain are the fastest-growing technologies in the network. The importance of the Human–Robot Systems cluster has also increased since 2015. Augmented and virtual reality were some of the key drivers of this growth. Additionally, the importance of other terms related to robotic systems, cobots, and autonomous robots grew in recent years.

Cloud Platforms and CAM are the two clusters whose importance has most declined in the last decade. Within the Cloud Platforms cluster, the OPC UA protocol gained the most importance. Among the smaller technologies, edge and fog computing did not play a role until 2018 but since then have shown strong growth in centrality measures. Technologies related to the CAM cluster, such as computerized numerical control (CNC), computer-aided design (CAD), and CAM, have rarely been emphasized in recent publications on Industry 4.0. Similarly, technologies in the Management Systems cluster are infrequently mentioned in recent Industry 4.0. articles. The most important technologies within the Management Systems cluster, such as manufacturing execution systems (MESs) and condition monitoring systems (CMSs), have been declining. Finally, the importance of the Additive Manufacturing cluster also decreased during our analysis period. While key technology terms such as 3D printing or additive manufacturing have recently gained importance and are part of many frameworks, the overall cluster (i.e., less relevant keywords) did not grow as much as keywords in other clusters.

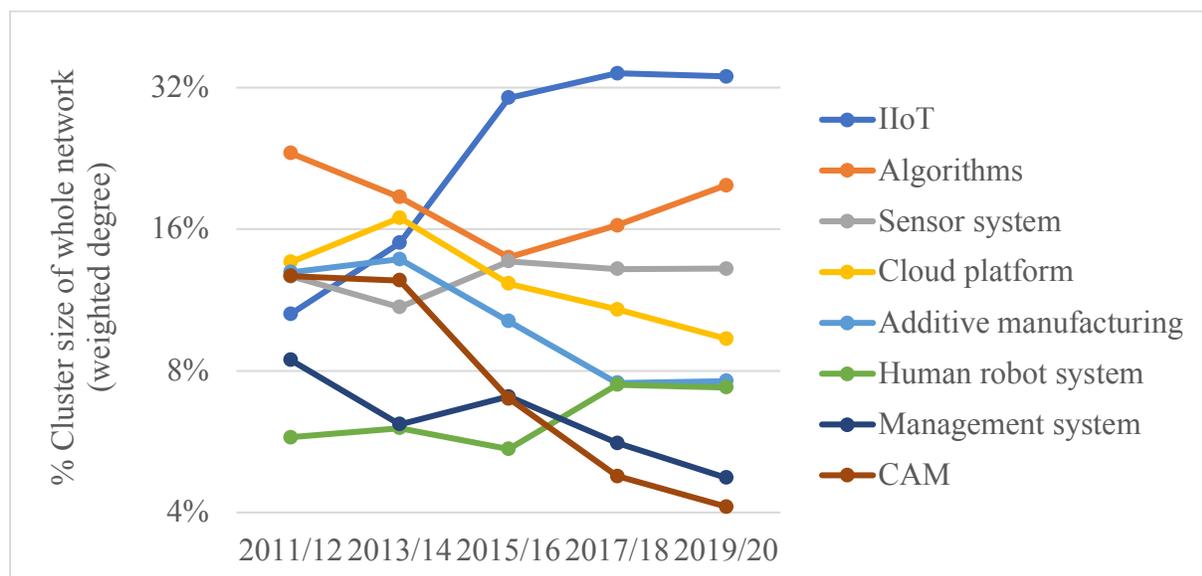

***Fig. 4.*** *Cluster sizes in recent years. We analyzed technology networks based only on the articles within each two-year period to calculate cluster sizes. For each period, we calculated the sum of the weighted degree of all technologies per cluster. The chart indicates the share of cluster degree of the overall size of the network (sum of*

*degrees of all nodes). IIoT refers to the Industrial Interent of Things cluster, and CAM to the computer aided manufacturing cluster. [1-column]*

The technology map (Fig. 3) provides some insights into the relations between clusters. Generally, the closer together two clusters are positioned on the map, the stronger the connection between them. However, the map has only two dimensions and thus cannot fully reflect the connections between all clusters. Therefore, we calculated the relative importance (RI) between all clusters. RI indicates the number of links between two clusters (i.e., whether a technology in cluster A is connected to a technology in cluster B). The RI value also normalizes for both clusters' total sizes, as there are generally more connections to larger clusters. Fig. 5 visualizes cluster RI, with darker shades indicating a higher score. For example, even though the IIoT cluster has strong connections to all clusters by total number, its RI is high only for the Sensor Systems and Cloud Platforms clusters. This indicates a particularly strong connection to those clusters relative to IIoT, Sensor Systems, and Cloud Platform's total cluster sizes. There is high RI among Additive Manufacturing, CAM, and Human–Robot Systems, indicating a strong relationship between these clusters. Human–Robot Systems also has high RI for Sensor Systems and Algorithms and low RI for Management Systems. Finally, the Management Systems cluster has high RI for Cloud Platforms and Additive Manufacturing.

|  | Size | AL | MS | AM | CAM | HRS | SE | IIoT | CP |
|---|---|---|---|---|---|---|---|---|---|
| AL | 5,131 |  |  |  |  |  |  |  |  |
| MS | 1,597 |  |  |  |  |  |  |  |  |
| AM | 2,415 |  |  |  |  |  |  |  |  |
| CAM | 1,550 |  |  |  |  |  |  |  |  |
| HRS | 2,040 |  |  |  |  |  |  |  |  |
| SE | 3,787 |  |  |  |  |  |  |  |  |
| IIoT | 9,254 |  |  |  |  |  |  |  |  |
| CP | 3,110 |  |  |  |  |  |  |  |  |

***Fig. 5.*** *Heatmap of relations between technology clusters. Numbers indicate the total number of links (measured in weighted degree), and the color code indicates the relative importance between clusters, with darker colors indicating greater importance. Cluster names: AL: Algorithm, MS: Management Systems, AM: Additive Manufacturing, CAM: Computer-aided Manufacturing, SE: Sensor Systems, IIoT: Industrial Internet of Things, CP: Cloud Platform. [1-column]*

Below, we examine cluster-bridging technologies in order to better understand how clusters are connected. These technologies, which have particularly strong connections to technologies outside their own cluster, are indicated in Table 2 along with their corresponding clusters. RFID, smart sensors, and wireless sensor networks (all in the Sensor Systems cluster) are technologies linking IIoT and Sensor Systems. Terms related to cloud computing and network communications—for example, OPC UA (Cloud Platforms cluster), Internet, and SDN (IIoT cluster)—form some of the links between IIoT and Cloud Platforms. The link between IIoT and Algorithms are reflected in the importance of big data analytics and AI (both in the Algorithms cluster) for both clusters. The Human–Robot Systems and Additive Manufacturing clusters have several bridging technologies, such as augmented reality and autonomous robots. Human–Robot Systems also has high relevance to Algorithms; some of the relevant cluster-bridging technologies include virtual reality and cobots (both associated with Human–Robot Systems) and AI (Algorithms). Management Systems is most strongly linked to Cloud Platforms. Product lifecycle management (PLM) and CMS are some of the terms with the highest relevance for both clusters. Cloud Platforms also overlaps with Algorithms; relevant technologies comprise cloud computing, simulation, and big data analytics.

## 5. DISCUSSION

We discuss the evolution of the Industry 4.0 landscape, building on technology clusters as defined in section **Error! Reference source not found.**. These clusters summarize several closely related technologies and help to discuss the technology landscape. Clusters provide suggestions for interpretation, not definite constructs, and different cluster analysis parameters can lead to different numbers of clusters. The present eight clusters offer a level of detail suitable for discussion and lead to homogeneous clusters. Beyond these clusters, our dataset also offers complementary, more detailed insights at the level of individual technologies.

### 5.1. Technology trends

In 2011, Algorithms, Cloud Platforms, and CAM were the most important clusters in the technology landscape. Together with Sensor Systems, Additive Manufacturing, and Management Systems, IIoT was among the less important technologies. Human–Robot Systems did not play a significant role. Our results show that a new paradigm evolved in 2015–2016, when IIoT became by far the most important cluster, followed by Algorithms. Management Systems, CAM, and Additive Manufacturing played only minor roles, with each representing less than 6% of technologies within the landscape. With most articles currently related to Industry 4.0 and smart manufacturing, it appears that cluster importance trends have stabilized since 2015–2016.

The IIoT cluster, including CPSs, is the largest cluster and forms the core of the Industry 4.0 landscape. The cluster's central position in the technology map indicates its overall relevance (in terms of frequency) and its relatedness to the other technology clusters. IIoT has a particular strong connection to the Algorithms, Cloud Platforms, and Sensor Systems clusters. This is in line with literature that describes the IoT and CPSs as the core of Industry 4.0 (Pereira & Romero, 2017). The Industry 4.0 working group (Kagermann et al., 2013) stated that "the fourth industrial revolution [is] based on Cyber-Physical systems" (p. 13), and Roblek et al. (2016) described the IoT as "central to the new industrial revolution" (p. 1). Zheng et al. (2020) found that the IoT was the most important enabling technology of Industry 4.0. Chiarello et al. (2018) and Rüßmann et al. (2015) defined the IIoT as an important element of Industry 4.0, but as one of several pillars rather than the core pillar. Our analysis shows that the term IIoT has recently become more central. Sisinni, Saifullah, Han, Jennehag, and Gidlund (2018) described the IIoT as the part of the IoT that

18focuses on manufacturing, stating that what is "usually addressed as IoT, could be better named as consumer IoT, as opposed to IIoT" (p. 4724), which focuses on manufacturing. The rising importance of IIoT within the technology network may reflect the understanding that, in the manufacturing context, IoT is referred to as IIoT, and future research should increase awareness of this differentiation.

The CAM cluster is the most declining cluster within the technology landscape. While the importance of CAD is growing, the decline of the cluster is driven by lower relevance of terms such as CNC, machine tool, and CAM. Although CNC machines may be a part of the future manufacturing landscape, research no longer explicitly addresses it. Culot et al. (2020) described CAD and CAM as "old" technology and did not include it as a dimension in their review. The CAM cluster also has high relative importance to the Management Systems cluster, which likewise contains "old" technologies, such as enterprise information system and ERP (Culot et al., 2020). Finally, this cluster has the strongest links to the Additive Manufacturing cluster, which is also a production-related cluster.

The Management Systems cluster—including MES, PLM, and CMS—has been declining in importance within the Industry 4.0 landscape. Technologies in this cluster are included in some Industry 4.0 frameworks (Dalenogare et al., 2018; Lasi et al., 2014), but many frameworks do not consider them as key technologies (Chiarello et al., 2018; Culot et al., 2020; Ghobakhloo, 2018; Zhong et al., 2017). Supply chain management is a growing technology term in this cluster, and ERP is only slightly declining. This suggests the growing importance of technologies that go beyond factory boundaries, whereas technologies relating to the factory itself (CMS, MES) are declining in importance. Yu, Xu, and Lu (2015) described integration beyond company boundaries as the particular focus of cloud manufacturing. However, integration also plays an important role in other concepts that describe, for example, horizontal and vertical integration as a key pillar of Industry 4.0 (A. G. Frank, Dalenogare, et al., 2019; Rüßmann et al., 2015). Saucedo-Martínez et al. (2018) even described it as the most important technology field of Industry 4.0.

The Cloud Platforms cluster has slightly decreased in importance within the manufacturing landscape but remains important. This role is in line with the literature, which frequently describes it as a core Industry 4.0 technology (A. G. Frank, Dalenogare, et al., 2019; Zheng et al., 2020). Further, many cloud technologies have become standard and therefore less research is focused on these technologies. For instance, Liao et al. (2017) noted that the OPC UA standard (a technology in the Cloud Platforms cluster) has become standard in machine-to-machine communications. The strong connection of the Cloud Platforms cluster to the IIoT cluster underlines the high interdependency of the two. Cloud Platforms focuses on cloud computing, protocols, and platforms, which are an integral part of the IIoT, as they are related to IoT infrastructure, CPS, and devices. Boyes, Hallaq, Cunningham, and Watson (2018), for example, described cloud computing as an (optional) part of the IIoT.

The literature increasingly highlights challenges related to cloud computing systems in the context of the IIoT. The challenges of large, highly connected, and centrally controlled networks include high central data accumulation, reliability issues, and high latency (Pan & McElhannon, 2018; Wang, Luo, Jia, Liu, & Xie, 2020). Some of these challenges may be overcome by decentralizing some cloud capabilities into local data centers and intelligent devices (Georgakopoulos, Jayaraman, Fazia, Villari, & Ranjan, 2016). Alcácer and Cruz-Machado (2019) suggested that Industry 4.0 "will be the extinction of the centralized applications used in common manufacturing environments" (p. 915), and Nakayama et al. (2020) described this decentralization as the main difference between Industry 3.0 and 4.0. These decentralized systems are called edge or fog computing—terms that have recently grown significantly in the technology map but are still missing from many frameworks and play only a minor role in the Industry 4.0 literature (Culot et al., 2020; Kipper et al., 2020; Liao et al., 2017; Muhuri et al., 2019; Osterrieder, Budde, & Friedli, 2020; Rüßmann et al., 2015). However, some terms that also describe distributed intelligence—such as smart sensor, AGV, and autonomous robot—are more present in the literature. Future frameworks could emphasize the distributed nature of networks to facilitate overcoming challenges of network complexity.

Blockchain is among the fastest-growing technologies in our analysis. The technology offers potential benefits related to security, privacy, resilience, and reliability in an increasingly interconnected manufacturing landscape (Lee, Azamfar, & Singh, 2019). Its potential benefits are particularly relevant for decentralized (edge or fog) systems—for example, enabling reliable communication between devices without the need to use a central server for security or reliability. While many Industry 4.0 frameworks (Chen, 2017; A. G. Frank, Dalenogare, et al., 2019; Kagermann et al., 2013; Rüßmann et al., 2015) did not refer to blockchain technology, future frameworks should follow some recent articles (Bai et al., 2020; Gaiardelli et al., 2021; Zheng et al., 2020) and account for its growing importance within their frameworks.

In general, the Algorithms cluster has been an important element of the manufacturing landscape throughout the timeline of this analysis. Traditional analytics-related terms—such as analytics, genetic algorithm, or multi-agent system—declined in importance in recent years. In contrast, AI, machine learning, and big data analytics are among the fastest-growing terms. While AI already has a high centrality score, its recent strong growth in centrality suggests that it is becoming an increasingly central element of the manufacturing landscape. On the one hand, generally increasing awareness of AI could account for the strong recent AI growth (M. R. Frank, Wang, Cebrian, & Rahwan, 2019). On the other hand, this growth might reflect the increasing number of feasible AI use cases. Chui et al. (2018), for example, identified hundreds of AI use cases already implemented in companies around the world. The share of articles related to algorithm-focused frameworks (intelligent manufacturing, cloud manufacturing) has recently remained constant. This indicates the growing importance of the Algorithms cluster, particularly AI, driven by the growing presence of the topic across all Industry 4.0 concepts. If this trend continues, AI may become the core element of future Industry 4.0 landscapes.

20The Human–Robot Systems cluster has been the fastest-growing cluster aside from IIoT. This cluster describes an environment of intelligent robots, which are frequently described in the literature as using sensors to interact with users (Robla-Gomez et al., 2017) and learning from humans through gestures or speech (Du, Chen, Liu, Zhang, & Zhang, 2018). It is closely linked to Sensor Systems, IIoT, and Algorithms. One of the fastest-growing technologies in this cluster is augmented reality, which can support human workers in future workstations by overlaying machine information with the real-world environment. This can be used to provide assembly instructions for customized products (Mourtzis, Zogopoulos, & Xanthi, 2019), train employees (Longo, Nicoletti, & Padovano, 2017), support teleoperated industrial assembly tasks (Brizzi et al., 2018), or conduct quality control of manufactured parts (Butt, 2020). Through its various capabilities, augmented reality may become a key enabling technology for augmenting workers in a future smart workplace (A. G. Frank, Dalenogare, et al., 2019; Longo et al., 2017) and experience strong growth in importance in the future (Masood & Egger, 2019).

The importance of the Additive Manufacturing cluster within the Industry 4.0 landscape has been declining since 2011 and currently represents a small share of keywords. This low share reflects the literature, where many articles do not consider additive manufacturing a key Industry 4.0 technology (Hermann et al., 2016; Strozzi et al., 2017). While Liao et al. (2017) and Muhuri et al. (2019) described additive manufacturing as a key Industry 4.0 technology, their reviews did not identify it as one of the most important terms. Similarly, Zheng et al. (2020) defined it as a key technology in their review but found that it played a role in few Industry 4.0 research articles. At the same time, additive manufacturing is included in many other frameworks (Culot et al., 2020; A. G. Frank, Mendes, Ayala, & Ghezzi, 2019) and is considered a critical aspect of Industry 4.0 (Kumar, 2018), with an impact across the product lifecycle (Butt, 2020). Two aspects may drive this decline in importance. First, the share of frameworks with a focus on the Additive Manufacturing cluster, such as digital and advanced manufacturing, is declining and thus related terms are declining as well. Second, additive manufacturing was present in the literature before the Industry 4.0 concept was introduced. After the introduction of this concept, IIoT, sensor systems, and human–robot systems became much more relevant in manufacturing research, and consequently the share of articles on additive manufacturing decreased.

Additive Manufacturing has strong connections to the Human–Robot Systems cluster. Some important bridging technologies are robotics and autonomous robots. These links suggest a vision wherein additive manufacturing is fully integrated with the manufacturing environment. This could lower complexity in production and decentralize production (Mehrpouya et al., 2019). However, the application of additive manufacturing in mass production remains limited in the near term (Roca, Vaishnav, Mendonça, & Morgan, 2017), particularly due to its low throughput (Korner et al., 2020; Mehrpouya et al., 2019). Dalenogare et al. (2018) found that companies currently see the benefits of additive manufacturing in terms of product-related (development, lead time, customization) rather than production-related aspects (costs, productivity, process control). Rapid prototyping has already been established in many companies (Mehrpouya et al., 2019), which is reflected in our analysis, where CAD (in the CAM cluster) has strong links to the Additive Manufacturing cluster. Augmented reality, another strong link between Human–Robot Systems and Additive Manufacturing, can also contribute to the product design process (Butt, 2020). With augmented reality and additive manufacturing, objects can be visualized in a real-life setting and directly manufactured (Mourtzis, Papakostas, Mavrikios, Makris, & Alexopoulos, 2015). The high customization and flexibility of additive manufacturing can be valuable for maintenance activities (Butt, 2020). Ceruti, Marzocca, Liverani, and Bil (2019) describe a use case for additive manufacturing wherein it is used to produce spare parts in aerospace, with augmented reality facilitating maintenance work.

## 6. CONCLUSIONS

Over the last 10 years, Industry 4.0 has evolved as a key research topic in management, production, and operations research, but it is still not clearly defined. In this paper, we clarified the concept and boundaries of Industry 4.0, identified the most relevant technology trends, and analyzed its evolution. By creating a technology map, we built on network analysis to isolate technology clusters and identify relations between them. We use NLP and network analysis to review the technology trends described in more than 14,000 articles referring to Industry 4.0 and related concepts.

With the introduction of Industry 4.0, the IIoT has become a central aspect of the technology map. The IIoT reaches beyond company boundaries, thereby minimizing the role of local production management and control systems. A key aspect of the IIoT is the cloud. Due to the complexity of highly interconnected networks, distributed systems like edge computing and smart sensors and devices will play more important roles in the future. Blockchain technology may be key in enabling the reliability of these systems. These decentralized connected systems enable faster adoption of AI in manufacturing, which has been strongly growing in importance in the overall technology landscape during recent years. These intelligent systems will facilitate human–robot systems, including augmented reality, enabling workers to interact with robots naturally. Due to its high potential to simplify and decentralize production, additive manufacturing may also evolve in relevance from product development, small batch production, and maintenance to mainstream production once productivity further improves.

Industry 4.0, as used in the literature, is frequently still defined around the IIoT and CPSs. We suggest accounting for the increasingly central role of AI when defining the fourth industrial revolution. The Industry 4.0 working group (Kagermann et al., 2013) suggested that the fourth industrial revolution is "based on Cyber-Physical Systems" (p. 13). This is a narrow definition compared to how the group describe previous industrial revolutions. For example, they state that the second industrial revolution "follows [the] introduction of electrically-powered mass production based on the division of labor" (p. 13). This description does not limit the revolution to the conveyor belt (a key component of that revolution) but rather its broader—revolutionary— impact: the division of labor. Similarly, the fourth industrial revolution should not be narrowly conceived in terms of IoT-based factories. Instead, a definition could recognize the revolutionary use of AI as a core driver and more natural human–machine interactions as a new way of working across enterprise boundaries and along the product life cycle.

In addition to providing a better understanding of Industry 4.0, this work contributes to the general scientific literature by presenting a new approach for literature reviews. We extracted named entities from article abstracts using NLP, a method that allowed us to evaluate technology terms without relying on article keywords. Future researchers can apply these methods to different fields, and named entity recognition may be trained to recognize not only technologies but also any other concept, such as technical information, chemical formulas, and historical events. We then analyzed the technologies using network analysis. This analysis offers insights into technology clusters, relations, and trends, rather than looking only at word frequencies. While network analysis is commonly used to describe social interactions (social network analysis), we show that this method offers high potential for other fields, such as operations research.

Our work has some limitations and offers potential for future researchers to develop our approach. Our analysis describes the evolution of Industry 4.0 technologies. Future work could use additional databases to confirm these trends and gain additional insights. Including Clarivate Web of Science would add other articles to the analysis. Using other types of data, such as Industry 4.0–related patents, could also lead to additional insights and allow for additional analysis of the Industry 4.0 technology landscape (e.g., comparing trends in patents and scientific literature). Our machine learning–based NLP approach offers great flexibility and helped us identify a large range of technologies. With the field of NLP rapidly improving, future researchers may achieve even higher accuracy in identifying technology terms with novel algorithms and future releases of the tools used in our work. Finally, future researchers might build on the method introduced in this article to undertake additional analyses, such as reviewing the business impact of Industry 4.0 or conducting a long-term review of technology concepts and shifts in the landscape. This could further illuminate technological change and create early indicators for future technological shifts.

## ACKNOWLEDGEMENTS

The authors thank the Studienstiftung des Deutschen Volkes and University Lisbon UIDP / 50009/2020 for financial support; MIT Portugal, who supported the visiting stay at the Massachusetts Institute of Technology; and Explosion AI, who granted access to the tool Prodigy for annotating named entities.

26

28NIST. (2014). Smart Manufacturing Operations Planning and Control. Retrieved August 1, 2019, from https://www.nist.gov/system/files/documents/2017/05/09/FY2014_SMOPAC_ProgramPlan.pdf

Opsahl, T., Agneessens, F., & Skvoretz, J. (2010). Node centrality in weighted networks : Generalizing degree and shortest paths. *Social Networks*, *32*(3), 245–251. https://doi.org/10.1016/j.socnet.2010.03.006

Osterrieder, P., Budde, L., & Friedli, T. (2020). The smart factory as a key construct of industry 4.0: A systematic literature review. *International Journal of Production Economics*, *221*, 107476. https://doi.org/10.1016/j.ijpe.2019.08.011

Pan, J., & McElhannon, J. (2018). Future Edge Cloud and Edge Computing for Internet of Things Applications. *IEEE Internet of Things Journal*, *5*(1), 439–449. https://doi.org/10.1109/JIOT.2017.2767608

Paritala, P. K., Manchikatla, S., & Yarlagadda, P. K. D. V. (2017). Digital Manufacturing- Applications Past, Current, and Future Trends. *Procedia Engineering*, *174*, 982–991. https://doi.org/10.1016/j.proeng.2017.01.250

Pereira, A. C., & Romero, F. (2017). A review of the meanings and the implications of the Industry 4.0 concept. *Procedia Manufacturing*, *13*, 1206–1214. https://doi.org/10.1016/j.promfg.2017.09.032

Pirola, F., Boucher, X., Wiesner, S., & Pezzotta, G. (2020). Digital technologies in product-service systems: a literature review and a research agenda. *Computers in Industry*, *123*, 103301. https://doi.org/10.1016/j.compind.2020.103301

Prabhakaran, T., Lathabai, H. H., & Changat, M. (2015). Detection of paradigm shifts and emerging fields using scientific network: A case study of Information Technology for Engineering. *Technological Forecasting & Social Change*, *91*, 124–145. https://doi.org/10.1016/j.techfore.2014.02.003

Robla-Gomez, S., Becerra, V. M., Llata, J. R., Gonzalez-Sarabia, E., Torre-Ferrero, C., & Perez-Oria, J. (2017). Working Together: A Review on Safe Human-Robot Collaboration in Industrial Environments. *IEEE Access*, *5*, 26754–26773. https://doi.org/10.1109/ACCESS.2017.2773127

Roblek, V., Meško, M., & Krapež, A. (2016). A Complex View of Industry 4.0. *SAGE Open*, *6*(2). https://doi.org/10.1177/2158244016653987

Roca, J. B., Vaishnav, P., Mendonça, J., & Morgan, M. G. (2017). Getting past the hype about 3-D printing. *MIT Sloan Management Review*, *58*(3), 57–62. Retrieved from http://mitsmr.com/2l0tcXn

Rüßmann, M., Lorenz, M., Gerbert, P., Waldner, M., Justus, J., Engel, P., & Harnisch, M. (2015). *Industry 4.0: The Future of Productivity and Growth in Manufacturing Industries*. *The Boston Consulting Group*.

Santos, C., Mehrsai, A., Barros, A. C., Araújo, M., & Ares, E. (2017). Towards Industry 4.0: an overview of European strategic roadmaps. *Procedia Manufacturing*, *13*, 972–979.

31# APPENDIX: DETAILED DESCRIPTION OF NAMED ENTITY RECOGNITION FOR EXTRACTION OF TECHNOLOGY TERMS

Technology terms can comprise various words and are frequently presented in lists, such as "flexible and reconfigurable manufacturing systems." Standard NER techniques cannot be used to identify both "flexible manufacturing systems" and "reconfigurable manufacturing systems." Therefore, technology term heads—in this example, "systems"—are extracted as a first step. In the second step, the associated descriptive words—in this case, "flexible manufacturing" and "reconfigurable manufacturing"—are identified through dependency parsing and added to the headword.

**Identifying technology terms**

For NER, we used the open source NLP tool spaCy (Honnibal & Montani, 2017), which provides the highest overall accuracy compared to state-of-the-art NLP tools (Al Omran & Treude, 2017; Jiang, Banchs, & Li, 2016). Language processing consists of two tasks. First, the neural network is trained to recognize technology heads. Second, the technology heads, including descriptive words, are extracted from the text.

The tool Prodigy[§] was used to annotate technology terms for training the neural network. Prodigy selects phrases for annotation that are expected to have the highest impact on overall accuracy. Overall, 1,518 manual annotations were used to train spaCy's large language model and 454 annotations were used to evaluate the model for an accuracy of 78%. After 968 annotations, accuracy was already at 77%, indicating that additional annotations would not lead to improved accuracy.

The accuracy (precision) rate of 78% indicates that most technologies were correctly identified. Evaluation of recall (fraction of technologies identified) indicates similar accuracy (77%). However, NER can generally reach an accuracy of up to 85% (Strubell, Verga, Belanger, & McCallum, 2017). Therefore, future researchers could try to increase accuracy through, for example, improved preprocessing or more advanced neural modeling techniques (e.g., training the neural network with the recently introduced method BERT; Devlin, Chang, Kenton, & Toutanova, 2018). On the other hand, depending on the complexity of a named entity, the boundaries may be lower. With our 1,518 manual annotations, we exceeded the precision rate of Anick et al. (2014), who reached a precision rate of 63% for English technology terms based on 3,700 manual annotations.

**Extracting technology terms**

---

[§] https://prodi.gy/

In the second step, spaCy's large neural network model[**] identified describing words through dependency parsing and saved the whole entities. All words related to the headword via specific dependencies were added to the entity. Relevant dependency types, as described by Marneffe et al. (2014), include adjectival modifiers (amod), compounds, noun phrases as adverbial modifiers (npadvmod), and nominal modifiers (nmod). Fig. 6 provides an example of dependency types. Further, parsing rules were introduced to identify abbreviations of technologies (appos) and lists of technologies (conj).

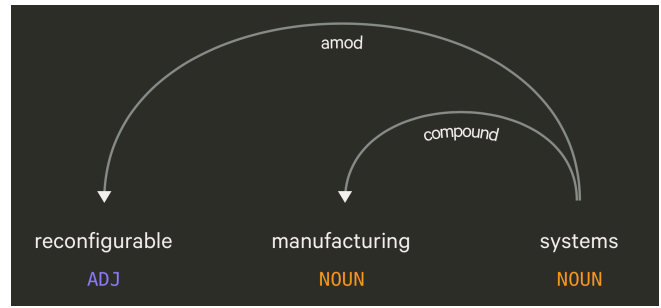

***Fig. 6.*** *Example of word dependencies displayed with the dependency visualizer displaCy from Explosion AI[††]. [1-column]*

**Post-processing**

A post-processing step was introduced to increase the quality of the results. First, words were lowercased and lemmatized using the natural language toolkit (NLTK) and the lemmatizer based on WordNet (Miller, Beckwith, Fellbaum, & August, 1990). This step avoids double counting of duplicate entities (e.g., changing "3D Printers" to "3d printer"). Second, leading words, which are commonly identified by the dependency parser but do not add to the specification of the named entity, were removed. These words include, for example, "novel," "expensive," and "first," as in "novel 3D printer." Third, we used a blacklist to remove frequently identified false positive technology terms. This list was mainly created based on a manual review of the most common technology terms. For example, if the algorithm identified "Smart Manufacturing" as a technology term, we would exclude it, as it was one of our concept search terms. This manual step helped avoid bias in technology clusters through links to non-technology terms. Additional rules were introduced to account for very specific technology terms. In the phrase "Internet of Things," "Internet" forms the headword; therefore, we introduced a rule to include "of Things" to technology term headwords.

---

[**] SpaCy models: en_core_web_lg-2.2.5 retrieved from https://github.com/explosion/spacy-models/releases/tag/en_core_web_lg-2.2.5, accessed on 10.01.2020.

[††] Image retrieved on Nov. 4, 2018 from https://explosion.ai/demos/displacy